# A new hazard event classification model
# via deep learning and multifractal

**Zhenhua Wang[1], Bin Wang[2], Ming Ren[1]\*, Dong Gao[2]**

[1]School of information resource management, Renmin University of China, Beijing, 100029, China

[2]College of Information Science and Technology, Beijing University of Chemical Technology, Beijing, 10029, China

zhenhua.wang@ruc.edu.cn; 2022400214@mail.buct.edu.cn; renm@ruc.edu.cn; gaodong@mail.buct.edu.cn

***Abstract***: Hazard and operability analysis (HAZOP) is the paradigm of industrial safety that can reveal the hazards of process from its node deviations, consequences, causes, measures and suggestions, and such hazards can be considered as hazard events (HaE). The classification research on HaE has much irreplaceable pragmatic values. In this paper, we present a novel **deep learning** model termed DLF through multi**f**ractal to explore HaE classification where the motivation is that HaE can be naturally regarded as a kind of time series. Specifically, first HaE is vectorized to get HaE time series by employing BERT. Then, a new multifractal analysis method termed HmF-DFA is proposed to win HaE fractal series by analyzing HaE time series. Finally, a new hierarchical gating neural network (HGNN) is designed to process HaE fractal series to accomplish the classification of HaE from three aspects: severity, possibility and risk. We take HAZOP reports of 18 processes as cases, and launch the experiments on this basis. Results demonstrate that compared with other classifiers, DLF classifier performs better under metrics of precision, recall and F1-score, especially for the severity aspect. Also, HmF-DFA and HGNN effectively promote HaE classification. Our HaE classification system can serve application incentives to experts, engineers, employees, and other enterprises. We hope our research can contribute added support to the daily practice in industrial safety.

### *NOMENCLATURE*

Notations and Observations

| | |
|---|---|
| HAZOP | Hazard and operability analysis |
| HaE | Hazard event |
| DLF | The proposed hazard event classification model |
| HGNN | The proposed hierarchical gating neural network |
| HTS | Hazard event time series |
| HFS | Hazard event fractal series |
| mF-DFA | Multifractal detrended fluctuation analysis |
| HmF-DFA | The proposed variant of multifractal detrended fluctuation analysis |
| BERT | Bidirectional encoder representations from transformers |
| CNN | Convolutional neural network |
| BiLSTM | Bidirectional long short-term memory |
| FC | Fully connected neural network |
| GMBC | The proposed gating mechanism |

## *1. INTRODUCTION*

The prosperity and advance of industry are of great significance to the development of the national volume. Yet, what we must to be wary of is that the complex process system involves miscellaneous materials and large-scale equipment, and frequently operates in extreme environments such as high temperature and high pressure, which is very easy to cause a series of safety accidents, resulting in casualties, economic losses and other disastrous consequences. Fortunately, hazard and operability analysis (HAZOP) can solve the dilemma (Kletz, 2018; Standard, 2001). HAZOP is a general and effective safety technique for revealing hazards, which takes each deviation (usually abnormal states such as countercurrent and high liquid level) of the system node (such as specific equipment) as the breakthrough point, retroactively traces its causes, infers its consequences in forward, and gives practical suggestions and measures. The results are reflected in HAZOP report, see Table 1 for two brief results in English format. (Wang(a) et al., 2022; Wang et al., 2022; Zhao et al., 2022; Wang et al., 2021; Peng et al., 2021; Feng et al., 2021).

Table 1: Two examples of results in the HAZOP report.

| Process node | Deviation | Consequences | Causes | Measures | Suggestions |
|---|---|---|---|---|---|
| Release gas compressor knockout drum. | Pressure is too high. | Resulting in equipment damage, sealing point leakage, and fire explosion. | Caused by the failure of the pressure controller PIC0801. | Add a PSA013 pressure safety valve and a PI0803 pressure indicator. | It is suggested to cancel PIC0801 split-range control to realize single-loop control of tank top pressure regulation. |
| Decarburization gas knockout drum. | Temperature is too low. | Resulting in freezing, blocking and leakage of process pipelines. | Caused by the low ambient temperature. | Set the insulation and heat tracing. | It is suggested to set temperature alarm at process pipeline flow. |

Currently, HAZOP has been popularized in various industries, such as human practice, petrochemical industry, food industry, biomass supply process, fusion engineering test and heat treatment system, etc. (Fattor and Vieira, 2019; Marhavilas et al., 2021; Khan et al., 2022; Cheraghi, et al., 2019; Wang et al., 2022; Lim et al., 2021; Lim et al., 2018; Wang et al.,2019; Patle et al., 2021; Hrjs et al., 2022). What's more, for some major countries, such as China, HAZOP has a considerable status, it must be executed for each process that to be put into released (China General Administration of work safety, 2013), and a large number of national policies declare its irreplaceability (Emergency Management Department of the people's Republic of China, 2019; Emergency Management Department of the people's Republic of China, 2020; Emergency Management Department of the people's Republic of China, 2022). Thus, HAZOP is a safety paradigm, and the research on HAZOP report is quite promising and necessary.

Different from the hazards in other fields, each result (each row of the table is organized into a result) in HAZOP reports corresponds to each hazard revealed by HAZOP under three aspects: severity, possibility and risk, which has the following particularities: (1) it is a complete statement with relatively long length, (2) formally it is the evolution of an event, and (3) it has its own logic and semantics. Therefore, we consider this hazard as hazard event (HaE), see two HaE in Table 1. The classification of HaE can promote the intelligent development of industrial safety, for example, it can support experts to explore the process that has not been put into production and assist engineers to launch the determination of decision-making for safety precautions, etc. Researches on relevant hazard classification in other fields also indirectly prove the importance and practicality of our work (Tian et al., 2022; Tanguy et al., 2016; Zhong et al., 2020; Pedrayes et al., 2022; Silva et al., 2022; Ferreira et al., 2019; Jahani, 2017; Júnior et al., 2016; Ouyang et al. 2022; Zheng et al., 2022).

At present, there is no complete classification research on HaE, and only one work with some relevance comes from Feng et al (2021). They used deep learning models BERT, BiLSTM and Attention to classify the consequences in HaE on severity aspect, and the experimental case data are three HAZOP reports of diesel hydrogenation, residue hydrogenation and pre-hydrogenation units. However, their work has obvious deficiencies: (1) it fails to consider the possibility aspect and risk aspect, only the severity aspect, (2) the research object is not the HaE, but the consequence in HaE; which weaken the power of HAZOP in redevelopment.

In view of the above, in this paper, we conduct the classification study on HaE and propose a new HaE classification model termed DLF via **d**eep **l**earning with multi**f**ractal, which mainly has two motivations.

1. Wang et al. (2022), pointed out that HaE satisfies the causality of its deviation, cause, consequence, measure and suggestion, and HaE can evolve and change over time, since what it undertakes is the logic that hazard is triggered under actual conditions. So, we can naturally regard HaE as a kind of time series.

2. Encouraged by fractal theory (Andres et al., 2020), the introduction of multifractal analysis is expected to improve the model's understanding of HaE, since HaE has semantics and logic, and can meet certain latent spatial distribution and has self-similarity. Fortunately, some representative studies confirmed the effectiveness and superiority of fractal theory in processing time / text series (Wang and Shao, 2020; Wang et al., 2020; Wen and Cheong, 2021).

Specifically, the proposed DLF first manipulates BERT for further pre-training to vectorize HaE. Then, through the proposed new multifractal method termed HmF-DFA based on multifractal detrended fluctuation analysis, DLF calculates HaE vector conditioned on the time series to court HaE fractal series dependent on Hurst exponent. Finally, DLF designs a hierarchical gating neural network (HGNN) to investigate HaE fractal series to accomplish the classification of HaE from severity aspect, possibility aspect and risk aspect. We take HAZOP report of 18 processes as cases. On this basis, we collect HaE for experiments to evaluate DLF classifier. The experimental results demonstrate that DLF classifier is gratifying and promising, HmF-DFA and HGNN are effective, and the idea of extending multifractal to HaE has received approvals and supports.

Our HaE classification system can serve and heighten HAZOP intelligently and orderly, which can support expert teams to explore new processes, assist engineers to manage emergencies, and facilitate employees to perform routine maintenance. Besides, it can guide other relevant enterprises to conduct safety analysis on industrial processes.

Our research can provide a new posture and inspiration for other researchers who are committed to the intelligent development and autonomous perception of industrial safety. The main highlights are as follow.

1. We contribute a novel deep learning model with multifractal termed DLF for HaE classification.

2. We propose a variant of multifractal detrended fluctuation analysis termed HmF-DFA.

3. We design a new hierarchical gating neural network termed HGNN.

4. Experiments based on cases of multiple processes prove the effectiveness of DLF, HmF-DFA and HGNN.

5. HaE classification system can bring application incentives to experts, engineers, employees, and other enterprises.

Section 2 mainly introduces multifractal analysis and hazard classification. Section 3 completely illustrates our DLF classifier. Section 4 presents cases. Section 5 elaborates experiments and analyzes the results. Section 6 records the applications of HaE classification system. Section 7 starts the discussion. Section 8 presents the conclusion.

## 2. RELATED WORK

### 2.1. Multifractal analysis

Multifractal analysis is a cutting-edge concept and method. In a broad sense, reveals the self-similarity between the local and the whole in the objective, and the latter can be embodied and deepened through the understanding of the former. People have the opportunity to deeply understand the potential mechanism of complex systems by exploiting fractals that is ubiquitous in natural science and social science. At present, multifractal analysis is being applied and explored in many fields and tasks (Mwema et al., 2020; Nguyen et al., 2022; Lancaster et al., 2021; Li et al., 2022; Pavlov et al., 2020; Clemente and Cornaro, 2022; Wang(b) et al., 2021; Xu et al., 2021).

Multifractal detrended fluctuation analysis (mF-DFA) is a critical and prevalent method for exploring the fractal properties hidden in various time series and quasi-time series, such as: customer review sentiment analysis (Zhang et al., 2022), gear fault diagnosis (Lin et al., 2021), comparative efficiency of green and conventional bonds (Naeem et al., 2021), temperature across Spain (Gómez et al., 2021), electroencephalogram cognition (Gaurav et al., 2021), human translation (Najafi et al., 2022), long-term memory retrieval (Kubota and Zouridakis, 2022), toll-free calls (Gui et al., 2021), agricultural image processing (Tang et al., 2020), etc.

In language, under the influence of grammar, the interaction between words forms specific meanings, which conveys the will, feelings and thoughts of human beings and society. The meaning of some text fragments can often express that of the whole text, and may be more accurate, concise and refined, which reveals the universal connection in language from a specific perspective. By discovering the fractal in words, language itself or some social behaviors can be more profound. Menzerath-Altmann law (Hou et al., 2017) can provide additional support for multifractal in language. It indicates that the longer a language structure is, the shorter its components are, that is, the part length is a function of the structure length. Another reminder is Hurst exponent (Mielniczuk and Wojdyłło, 2007), which reflects the autocorrelation of time series, especially the long-term trend and memory hidden in the series.

Naturally, we follow and vary mF-DFA to guide the fractal in the language carried by the text to bring extra surprise to the HaE classification. A representative study about emotion analysis (Zhang et al., 2022) provides a reference, the input is the time series formed by vectoring hotel reviews through Word2vec, the output is the generalized Hurst exponent that to be processed by the support vector machine.

### 2.2. Hazard classification

The classification research on hazards is essential to perfect industry assessment, promote safety management, and help care for employees, which reflects and highlights the humanitarian brilliance of enterprises and researchers (Tian et al., 2022; Tanguy et al., 2016; Zhong et al., 2020; Pedrayes et al., 2022; Silva et al., 2022; Ferreira et al., 2019; Barber et al., 2017; Stiernstroem et al., 2016; Ouyang et al. 2022; Zheng et al., 2022). So far, with the emergence and development of deep learning and natural language processing, extensive related work in different fields has arrived. For instance, Zhong et al. (2020), proposed a CNN-based classification model from subway hazard reports to prevent accidents. Jahani (2019) constructed a hazard classification feedforward neural network based on the fault of platanus orientalis, which is used as an environmental decision support system for reducing the risk of tree collapse. Fang et al. (2020), studied the BERT-guided classification of near misses in the construction industry to understand how to mitigate and control the hazards on the construction site. Moreover, LSTM-based classification model of accident causes for chemical hazards (Jing et al., 2022), CNN-based hazard behavior classification of mobile scaffold (Khan et al., 2021), LSTM-based safety-hazard classification for large scale construction project (Tian et al., 2022), classification study of aviation accident reports under n-gram language model (Tanguy et al., 2016), prediction of earth fission hazards by SVM classification (Choubin et al., 2019), fire hazard level system based on machine learning under wild fire history (Júnior et al., 2022), etc. (Pedrayes et al., 2022; Jeon and Cai, 2021; Ishola and Valles, 2022), also, imply the authenticity and reality of our research. Obviously, deep learning algorithms, especially CNN and BiLSTM, provide a superb foundation.

Different from the above work, our research object is the hazard event (HaE) analyzed by HAZOP. HaE is measured in three aspects (Kletz, 2018):

(1) Severity aspect. According to the degree of damage (such as personal injury), HaE is classified into five levels, for example, level {#1-#5} indicates that the degree of personal injury is {negligible, slight, serious, disabled, dead}.

(2) Possibility aspect. HaE has five levels based on the frequency at which it is triggered, level {#1-#5} indicates that the occurrence of HaE is {none, one decade ago, within one decade, within five years, within two years}.

(3) Risk aspect. Based on the acceptability under existing prevention, HaE enjoys four levels, level {#1-#4} indicates that it is {acceptable, tolerable, further investigated, immediately rectified}.

For example, for the aspect of {severity, possibility, risk} of two HaE in Table 1, the first one is {3, 2, 2}, which presents that 1) it can cause serious personal injury, 2) it occurred once a decade ago, and 3) it is tolerable under the existing protection. The second one is {4, 1, 2}, which implies that it disables people, yet never appears, thus, it is tolerable.

Our work aims to establish a classification model to analyze HaE from these three aspects. As far as we know, our work is the first comprehensive study. In view of the inspiration from other hazard classifications and similar work, deep learning is naturally our choice. In order to improve the performance of the classifier, multifractal analysis is taken into consideration.

Note that HAZOP reports protected by property rights are confidential. To a certain extent, our work has brought progress to the transfer and sharing of knowledge.

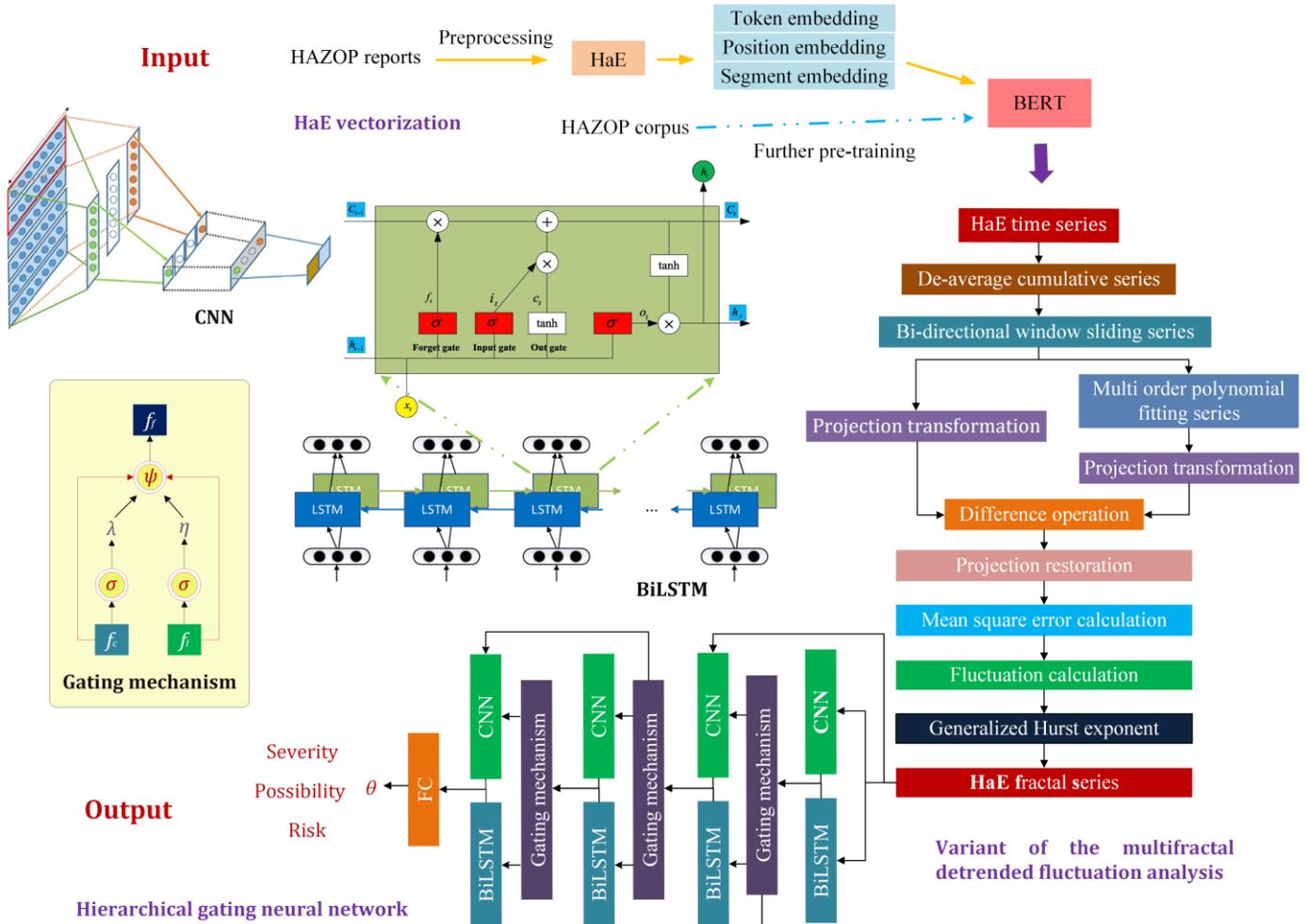

Fig.1: Architecture of DLF.

### 3. METHODOLOGY

We propose a new HaE classifier driven by deep learning and multifractal analysis, termed DLF, which can contribute to the intelligent development of industrial safety. This section is the whole procedure of DLF, see Fig.1. First, we vectorize the HaE from HAZOP reports through BERT to form HaE time series. Then, we treat HaE time series via the presented variant HmF-DFA of the multifractal detrended fluctuation analysis, to obtain HaE fractal series. Finally, we design a hierarchical gating neural network to investigate the HaE fractal series to recognize its classification. Details are as follows.

#### 3.1. HaE Vectorization

This section shows how to vectorize HaE. The input is HaE in text format and the output is HaE vector, see the upper part of Fig.1.

HaE is obtained from HAZOP reports through document preprocessing such as cleaning and arranging. Considering that the meaning of the HaE when the equipment is the subject is inconsistent with that when the equipment is the object, for example, the description when the "pressure indicating controller" is the subject often indicates the beginning or ongoing of a HaE, while the description when the equipment is the object is the measures and suggestions that confront with the hazard. In addition, HaE contains complicated terminologies and components of different processes, and there are also differences in their language styles, as well as the wording and phrasing.

Therefore, we vectorize HaE based on BERT that prospers industrial and social language understanding (Wang(c) et al., 2022; Melluso et al., 2022; Zheng et al., 2022; Yang et al., 2021; Zhu et al., 2021), which can well capture the semantic information. Note that BERT is trained from the general domain corpus, without the consideration of the professional domain corpus, and fails to enjoy the prior knowledge in the field of industrial safety. In order to make up for this deficiency, we carry out further pre-training on the HAZOP corpus through the same two self-supervision tasks as BERT, namely "masked language model" and "next sentence prediction" (Feng et al., 2021). In this way, the represented HaE vector is mapped by the elements in HaE dynamically in various contexts, which is more rational and appropriate.

Specifically, first, we preprocess the one-dimensional HaE text $W = \{w_1, w_2, \ldots, w_n\}$, add the "CLS" mark at the beginning of each HaE to establish its boundary, that is HaE = {CLS, $w_1, w_2, \ldots, w_n$}. Next, each processed HaE is segmented into a series of tokens. Tokens are indexed in the vocabulary provided by BERT to form the token embedding. The position embedding is formed according to predetermined the sine/cosine rules. Adjacent sentences are marked with 0 and 1 to form the segment embedding. Then, we straightforwardly pass the sum of the three embeddings into the decoder group of *Transformer* to generate the HaE vector with the dimension of 768 for each word. In this way, each HaE can be mapped into a two-dimensional matrix. The mean of the matrix is calculated and considered as a kind of time series, namely HaE time series (HTS) (Zhang et al., 2022).

The potential features contained in HTS are vague and convoluted, and it is not easy to directly distinguish the expected differences between its respective levels. In combination with the particularity of HaE (see Introduction section), and inspired by multifractal methods that have made great achievements in various fields (Mwema et al., 2020; Nguyen et al., 2022; Lancaster., 2021; Li et al., 2022; Pavlov et al., 2020; Clemente et al., 2022; Wang(b) et al., 2021; Liu et al., 2020; Zhang et al., 2022), we introduce the popular multifractal detrended fluctuation analysis (mF-DFA) (Miloş et al., 2020) to HTS from the perspective of language. Considering the severity aspect, possibility aspect and risk aspect, the distribution of their levels is unbalanced (as seen in Table 3), since the higher the level, the rarer the HaE, while the lower the level, the easier the HaE is usually triggered, which may bring restrictions to the work of mF-DFA. To alleviate this situation and further improve the performance of the classifier, we propose a variant of mF-DFA for HaE termed HmF-DFA to explore the classification of different HaE by investigating the implicit self-similarity in nonlinear complex systems.

#### 3.2. HmF-DFA

As shown in the right part of Fig.1, HmF-DFA follows the procedure mF-DFA, studies the variances of the fluctuations by considering the increasing windows of HTS, to embrace the HaE fractal series from the HTS, as follows.

Given an HTS, $R = \{r_1, r_2, \ldots, r_l\}$ with length $l$, we find its profile series as Equ.1.

$$D = \sum_{i=1}^{k}(r_i - \mu_R), \ \ k = 1, 2, \ldots, l \tag{1}$$

That is, calculate the cumulative sum of subtracting the mean $\mu_R$ that meets:

$$\mu_R = \frac{1}{l}\sum_{i=1}^{l} r_i$$

Further, we change the time scale of detrended profile series $D$, and slide it bi-directionally in the way of the sliding window operation, see Equ.2, where, $[\cdot]$ stands for rounding function, $s$ is the length of the window. Therefore, we get $2W_s$ non-overlapping windows, and take $\varphi^n_t$ to mark the window with serial number $n$, where $0 < t < s$.

$$W_s = [l / s] \qquad (2)$$

We notice that each window has its own local trend $\psi^n_t$, and the concatenation of all local trends is related to the overall contour of $D$, which corresponds to the internal self-similarity between the two. The fitting of each $\psi^n_t$ is equipped with the least square method, i.e., for $s$ points of each window, fit $m$-order polynomial that is Equ.3.

$$\psi^n_t(k) = a_1 k^m + a_2 k^{m-1} + \cdots + a_k k + a_{i+1}, \quad k = 1, 2, ..., s, \quad m = 1, 2, ... \qquad (3)$$

The mF-DFA relishes the detrended HTS $\delta^n_t$ by calculating the difference between $\varphi^n_t$ and $\psi^n_t$. Yet, the unbalanced distribution of levels under three aspects may cause additional conflict to the $\delta^n_t$, and the difference between the profile series and its local trend may have undesired changes among different windows. To alleviate this condition, we propose HmF-DFA that can provide an optimization extension. Specifically, first, we project $\varphi^n_t$ and $\psi^n_t$ onto the plane where Equ.4 is located, one consideration is that $Sigmoid$ function family has a certain degree of data centralization effect (Minai and Williams, 1993), and the other is that $\alpha$ is a scaling index used to homogenize the output, when $\alpha$ approaches 0, Equ.4 degenerates into a linear function that can hold the original input distribution. Then we perform the difference value between the projected $\varphi^n_t$ and $\psi^n_t$ over the $Sigmoid$ plane. Finally, we restore the difference value through the inverse function of Equ.4 to accept the detrended HTS $\delta^n_t$ under HmF-DFA. See Equ.5, where, $\sigma*(\alpha, x)$ is the inverse function of $\sigma(\alpha, x)$, $\alpha$ is 0.4 empirically and experimentally.

$$\sigma(\alpha, x) = sigmoid(\alpha x) = 1 / (1 + e^{-\alpha x}) \qquad (4)$$

$$\delta^n_t = \sigma^*(\alpha, [\sigma(\alpha, \varphi^n_t) - \sigma(\alpha, \psi^n_t)]) \qquad (5)$$

The long-range correlation under the representation and behavior of HaE can be revealed by alleviating the non-stationarity in HTS. We calculate the local variance $\sigma^2(s, n)$ of the detrended HTS dominated by the window size $s$ variable and window series number $n$ variable, see Equ.6. For $2W_s$ windows, we calculate the mean value of $\sigma^2(s, n)$ to get the $q$-order fluctuation function $F_q(s)$, see Equ.7, (Pavlov et al., 2020).

$$\sigma^2(s, n) = \frac{1}{s} \sum_{t=1}^{s} (\delta^n_t)^2 \qquad (6)$$

$$F_q(s) = \begin{cases} [\dfrac{1}{2W_s} \sum_{n=1}^{2W_s} \sigma^2(s, n)^{\frac{q}{2}}]^{\frac{1}{q}}, \ q \neq 0 \\ exp\{\dfrac{1}{4W_s} \sum_{n=1}^{2W_s} ln(\sigma^2(s, n))\}, \ q = 0 \end{cases} \qquad (7)$$

By setting window size $s$ to observe $F_q(s)$, obviously, with the increase of $s$, $F_q(s)$ rises in a power-law in the form of Equ.8 which depends on the fractal order $q$.

$$F_q(s) \propto s^{H(q)} \qquad (8)$$

We can obtain the generalized Hurst exponents $H(q)$ (when $q = 2$, $H(q)$ is the standard Hurst exponents) by fitting $F_q(s)$ with Boltzmann method (Meira et al., 2020), to form the **f**ractal **s**eries of **H**aE (HFS) with multifractal particularities that are reflected in the change of $H(q)$ with $q$.

Now, we have depicted the HFS. Next, we design a hierarchical gating neural network to investigate it.

### 3.3. Hierarchical gating neural network

The conceived hierarchical gating neural network (HGNN) considers BiLSTM and CNN since they are two popular neural networks for hazard analysis (see Section 2.2), and another idea is that HGNN designs a gating mechanism for feature fusion, see the lower of Fig.1.

We start with a brief introduction to BiLSTM (Wang(c) et al., 2022). BiLSTM is used to extract context features $f_c$ and consists of two bidirectional LSTMs, each LSTM has more forgotten gate $f_t$, input gate $i_t$ and output gate $o_t$ on the basis of RNN, see Equ.9-11 respectively, where, $W$ and $b$ represent respective input weight matrix and bias terms, $s$ indicates the hidden state, subscript $t$ is time step, and $HFS_t$ represents the input value at time $t$. BiLSTM bidirectionally splices the output $h_t$ of LSTM to form $f_c$.

$$f_t = sigmoid(W_f HFS_t + W_f s_{t-1} + b_f) \qquad (9)$$

$$i_t = sigmoid(W_i HFS_t + W_i s_{t-1} + b_i)$$
$$c'_t = \tanh(W_c HFS_t + W_c s_{t-1} + b_c) \qquad (10)$$
$$c_t = i_t \cdot c'_t + f_t \cdot c_{t-1}$$

$$o_t = sigmoid(W_o HFS_t + W_o h_{t-1} + b_o)$$
$$h_t = o_t \cdot \tanh(c_t) \qquad (11)$$

Next is CNN, which we apply to extract text local features $f_l$. Specifically, the convolution layer generates the feature graph $p$ through the convolution kernel with the size $d$ and the ReLU activation function, see Equ.12, where, $HFS_{k:k+d-1}$ is a sequence fragment of HFS indexes from $k$ to $k+d-1$, $n$ is the length of the HFS. We perform the *max*-pooling operation (equivalent to the *max* function) on the feature graph $p$ to get the prominent features, namely $f_l$.

$$p = \{ReLU(W_p \cdot HFS_{k:k+d-1} + b_p)\}_m \mid_{m=1}^{n-d+1} \qquad (12)$$

The fusion between different features can well stimulate the potential of features. For this reason, benefiting from the research of Gu et al. (2020), we design a new gating mechanism termed GMBC to fuse $f_c$ and $f_l$.

$$f_c = BiLSTM(HFS)$$
$$f_l = CNN(HFS) \qquad (13)$$

$$\lambda = sigmoid(W_c f_c + b_c)$$
$$\eta = sigmoid(W_l f_l + b_l) \qquad (14)$$

Specifically, first, HFS is encoded into $f_c$ and $f_l$ by BiLSTM and CNN, see Equ.13. Next, the two form their respective screening coefficients through *sigmoid* functions, i.e., $\lambda$ and $\eta$, see Equ.14, where, $W$ and $b$ are the learned weight terms and bias terms respectively. Then, Equ.15 calculates a feature assignment weight $\psi$ based on $\lambda$ and $\eta$. It can be analyzed that when $\eta \to 0$, $\psi \to \lambda$; when $\lambda \to 0$, $\psi \to \eta$. In these two cases, $\psi$ degenerates to the classical method (that is, *sigmoid* function mainly acts as the gating mechanism) of feature weight allocation (Deng et al, 2021). Thus, our $\psi$ is an optimization that can better weigh different features.

$$\psi = \eta \cdot (1 - (1 - \lambda)^2) + (1 - \eta) \cdot \lambda \qquad (15)$$

Finally, the fusion vector $f_f$ is formed by Equ.16, where, $\psi + \rho = 1$.

$$f_f = \rho \cdot f_c + \psi \cdot f_l \qquad (16)$$

After discussing GMBC, we expound the operation procedure of HGNN. Details are as follows.

The three feature vectors obtained by the first layer are shown in Equ.17, where the subscript indicates the layer, for example, the subscript "1" indicates the first layer.

$$f_{c-1}, f_{l-1} = BiLSTM(HFS), CNN(HFS)$$
$$f_{f-1} = GMBC(f_{c-1}, f_{l-1}) \qquad (17)$$

In the second layer, BiLSTM takes $f_{f-1}$ as the input, CNN takes the concatenation of $f_{f-1}$ and the original vector HFS as the input, and their outputs form $f_{f-2}$ via GMBC, see Equ.18.

$$f_{c-2}, f_{l-2} = BiLSTM(f_{f-1}), CNN(f_{f-1} \oplus HFS)$$
$$f_{f-2} = GMBC(f_{c-2}, f_{l-2}) \qquad (18)$$

In the third layer, the input of BiLSTM is the concatenation of $f_{f-1}$ and $f_{f-2}$, the input of CNN is $f_{f-2}$, and see Equ.19 for the three feature vectors.

$$f_{c-3}, f_{l-3} = BiLSTM(f_{f-1} \oplus f_{f-2}), CNN(f_{f-2})$$
$$f_{f-3} = GMBC(f_{c-3}, f_{l-3}) \qquad (19)$$

The operation of the fourth layer is similar to that of the second layer, see Equ.20, where, the fusion feature $f_{f-4}$ is the concatenation of $f_{c-4}$ and $f_{l-4}$.

$$f_{c-4}, f_{l-4} = BiLSTM(f_{f-3}), CNN(f_{f-3} \oplus f_{f-2})$$

$$f_{f-4} = f_{c-4} \oplus f_{l-4}$$

(20)

We leverage a fully connected neural network (FC) to map $f_{f-4}$ to $T$ with the dimension of category number, and then employ *softmax* to predict the classification $\theta$ from a discrete set of classes for $f_{f-4}$, see Equ.21.

$$\theta = argmax[softmax(T)]$$

$$= \begin{cases} Severity = \{level\,\#i\}\,|_{i=1}^{5} \\ Possibility = \{level\,\#i\}\,|_{i=1}^{5} \\ Risk = \{level\,\#i\}\,|_{i=1}^{4} \end{cases}$$

(21)

## 4. Cases

In order to guarantee the quality of the proposed DLF classifier and evaluate its effectiveness, we take 18 HAZOP reports as our research cases. These reports are developed and recorded for different processes under the arrangement between us and some well-established enterprises, which are of great significance in clean energy, environmental protection and sustainable development, etc. They are not only legal obligations, but also embodiments of social responsibility. The brief introduction is shown in Table 2, where, the system nodes subject to specific objects, the aspects refers to severity, possibility and risk in turn.

Table 2: Brief information about 18 HAZOP reports.

| Process | profile | Main nodes | Example of HaE | Aspects | Enterprises |
|---|---|---|---|---|---|
| 2.2 million T / a diesel hydrofining | Raw materials are straight-run diesel oil and straight-run kerosene from atmospheric and vacuum distillation unit, and coking gasoline and coking diesel oil from coking unit. These units mainly produce high-quality ultra-low sulfur refined diesel oil (with sulfur content no more than 10ppm), followed by naphtha and sulfur-rich gas. | Hydrogen mixing and heating of feed oil; Separator part; Corrosion inhibitor and water injection unit; New hydrogen compressor; Heating furnace fuel gas and air distribution system; Desulfurization hydrogen stripper; Product fractionating tower; Qualified diesel and naphtha outlet device; Low separation gas desulfurization tower; Tank collection; | The de-aerated water entering the middle section of R-5611101 is not available or the flow is too low, which makes its temperature is too high, and the reactor is extremely warm and leaking. In this case, the emergence of ignition source causes fire and explosion, resulting in poisoning and casualties. Execute TI0108-0113-AH. It is recommended to refine the response time of de-aerated water interruption. | {5, 5, 2} | Hengyi Petrochemical (http://www.hengyi.com/) |
| 4 million T / a indirect coal liquefaction | The raw coal reacts with oxygen and steam to completely gasify the coal, and the generated feed gas is transformed, desulfurized and decarburized to produce clean synthetic gas. The synthesis reaction of syngas takes place under the action of catalyst to generate hydrocarbons, which are further processed to produce gasoline, diesel, liquefied petroleum gas, etc. | Fischer Tropsch synthesis unit; Catalyst reduction unit; Wax filtration unit; Tail gas decarburization unit; Fine desulfurization unit; Synthetic water treatment unit; Liquid intermediate raw material tank farm unit; Low temperature oil washing unit; Deoxygenated water and condensate refining station unit. | The design of the bottom process of R-5611101 has changed, which makes the handover of its maintenance process abnormal, and the residue remains in the bottom treatment. In this case, renovation and overhaul may cause fire casualties. Fill the bottom of R-5611101 with ceramic balls. It is recommended to carry out handover activities during maintenance to record the log that the bottom wax oil cannot be discharged smoothly. | {3, 4, 2} | Shenhua Group (http://www.shenhuachina.com/) |
| 1.2 million T / a heavy oil catalytic cracking | The raw materials from the heating furnace or heat exchanger are atomized by steam and nozzle, and then sent to the transfer pipe to mix and react with the hot regeneration catalyst from the regenerator on the catalyst bed. The product enters the fractionation system and is then divided into liquefied gas, gasoline, light gas oil and clean oil. | Feed processing of raw oil; Heat exchanger system; Return to refining system; Steam generation system; Slurry treatment; Oil circulation system; Cooler system; Separator system and catalyst system. | The flow of slurry pump P-1209 is too small due to the fluctuation of slurry circulating volume. The temperature at the bottom of the fractionator is low, and in serious cases, the reaction feed is cut off, and the slurry pump P-1209 is evacuated. Standby pumps P1209B, FT20203 and FT20204. It is recommended that the pump P1209 be used for one operation and two standbys. | {2, 4, 1} | Sichuan Petrochemical (http://scsh.cnpc.com.cn/) |
| 30 thousand T / a desulfurization and sulfur recovery | Acid gas reacts with air in the combustion furnace to recover sulfur through condensation. The remaining gas is used for hydrogen sulfide and sulfur dioxide to produce sulfur, which is recovered through cooling. The tail gas is hydrogenated into $H_2S$ | Gas desulfurization system; Feed buffer system; Solvent regeneration and storage system; Acid and process gas reaction treatment system; Liquid sulfur generation link and tail gas treatment system. | The increase of gas phase volatilization leads to high temperature at the top of flash tower T-202, which makes the flow of foam rich liquid into the tower low. Use TG120 to monitor the feed temperature of rich | {2, 3, 2} | Liaoyang Petrochemical (http://lysh.cnpc.com.cn/) |
| | by hydrogen, which is used for further sulfur recovery in the acid gas combustion furnace. Waste gas is discharged into the atmosphere through the incinerator. | | solution. It is recommended to add tower top temperature indicator to control lean liquid flow on tower top by using tower top temperature | | |
| 600 thousand T / a light naphtha isomerization | The raffinate oil from the aromatics complex enters the feed buffer tank of the raffinate oil tower, and after pressurization, carries out heat exchange with the tower bottom liquid to remove the heavy raffinate oil. Some products are condensed for reflux, and some products are sent to the feed buffer tank of the deisopentanizer. The heavy raffinate is sent out of the unit after heat exchange with the feed. | Residual oil pumping system; Circulating feeding system; Deisopentane and its systems; Desulfurization system; Isomerization system. | The liquid level in the high-pressure vent tank is high due to the untimely discharge of liquid, which leads to the liquid is carried to flare system by pipeline; Turn down LV20401 and set slope to flare pipeline. It is recommended to cancel the alarm function of LAHH20401 and realize it by LI20402 circuit. | {2, 2, 4} | Hengyi Petrochemical (http://www.hengyi.com/) |
| 0.075 T / h ammonium nitrate to nitrous oxide | Melting pot → reactor → condenser → pre-asher → purification tower I (water washing) → purification tower III (alkali washing) → purification tower II (alkali washing) → purification tower IV (acid washing) → purification tower V (water washing) → N$_2$O compressor → water separator → molecular sieve dryer group → condensate liquefier → N$_2$O high-pressure storage tank → low-pressure tank. | Ammonium nitrate melting system; N$_2$O cooling; Washing and purification, compression, drying, buffering and liquefaction systems. | PIC0118 circuit fault causes high pressure of fuel gas tank V0102, resulting in the opening of PV0118, which may lead to incomplete combustion of F0101 natural gas; Execute PSV0153. It is recommended to monitor the oxygen content in furnace AT0101 online. | {2, 5, 3} | Universal Jinghui Technology (http://www.yjhqt.com/) |
| 500 m$^3$ / h water electrolysis hydrogen production | Under the action of direct current, water in the electrolytic cell is decomposed, hydrogen is separated from the cathode surface, oxygen is separated from the anode surface, and they are further purified to produce hydrogen and oxygen products. | Cooling solution; washing solution; Deoxidation and drying of hydrogen and oxygen; Hydrogen; Regeneration of dryer; Lye cooling cycle. | Due to the failure of LIC0101 circuit, the liquid level of steam drum V0103 is low, which causes the water of F0101 boiler to be burned dry. Monitor furnace temperature. It is recommended to conduct SIL grading analysis for this scenario to confirm whether to add safety instrumented system. | {3, 5, 3} | Universal Jinghui Technology (http://www.yjhqt.com/) |
| 100 m$^3$ / h formic acid to carbon monoxide | The formic acid is pumped to the dehydration reactor and co-heated with sulfuric acid for dehydration reaction. The generated CO enters the alkali scrubber to neutralize the acid saturated steam, and then enters the water scrubber to adjust the pH value to neutral. The neutralized saturated steam is cooled by the cooler and pumped into the circulating water system for circulation. CO enters the precooler for condensation and compression, and is purified by deoxidation, decarbonization and dehydration tower in turn, and then bottled after being pressurized by the compressor for three stages. | Dehydration reaction; Alkali washing, water washing, pressurization, deoxidation, decarburization and dehydration of CO; Regeneration of deoxidizer and drying tower; The product gas is pressurized and exported; Concentration, cooling and buffering of sulfuric acid. | The decompression of nitrogen hydrogen mixture makes the pressure of the regenerated deoxidizer high, resulting in valve failure, excessive opening, and leakage of the deaerator due to overpressure. Keep the regeneration gas outlet open. It is recommended to avoid using 16 bottles of high-pressure air simultaneously during regeneration, and but 1-2 bottles. | {2, 3, 2} | Universal Jinghui Technology (http://www.yjhqt.com/) |
| 800 m$^3$ / h natural gas hydrogen production | After pressurized desulfurization, natural gas and steam are cracked and reformed in a special reformer filled with catalyst to generate conversion gas of hydrogen, carbon dioxide and carbon monoxide. After some heat is recovered, CO content in the conversion gas is reduced through conversion, and the conversion gas is purified through pressure swing adsorption to obtain hydrogen gas. | Pressure regulating feeding; Natural gas desulfurization, reaction and conversion generate hydrogen; Fuel gas and furnace of heating furnace; Waste heat boiler; Deaerator system; Intermediate transformation of reforming gas. | Due to PIC1001 circuit fault, the pressure of oxygen separator 1003 is high, PV1001 is turned down, and the liquid level of oxygen separator drops. In serious cases, oxygen escapes from the oxygen phase pipeline into the hydrogen separator, causing an explosion. Add LSHH1003 hydrogen separator liquid level high interlock. It is recommended to make the liquid level alarm, control and interlocking independent. | {2, 5, 2} | Universal Jinghui Technology (http://www.yjhqt.com/) |
| 10 thousand T / a waste liquid desulfurization and sulfuric acid production | Pretreatment procedure: sulfur foam drying and XA sulfur burning. Acid making procedure: purification procedure, drying and absorption procedure and conversion section. | Trough system; Microporous filtration system; Flushing process; Water washing tower; Pickling tower; Clear solution concentration system; Steam condensation; Sulfur incinerator; Dynamic wave washing; Cooling tower; Gas drying and absorption system. | Due to excessive liquid discharge from the system, the liquid level in the underground storage tank V-2105 is high, and the liquid discharge is not timely, resulting in the overflow of waste liquid and environmental pollution. Drain P-2104 while delivering liquid. It is recommended to add liquid level monitoring to start and stop the pump. | {2, 3, 1} | Shengfa Coking Group (http://www.shengfajiaohua.com/) |
| 120 thousand T / a sulfur recovery | It mainly includes two series of 60 thousand T / a sulfur recovery unit and tail gas treatment unit, one series of solvent absorption and regeneration unit, sulfur forming unit and utility unit. | Claus sulfur recovery module; Sulfur forming module; Tail gas treatment module; Water supply module; Steam and condensate modules. | Due to the low temperature of the incinerator, the temperature of superheated and medium-pressure steam is low, resulting in burns when the high-temperature valve is manually vented. Discharge unqualified steam. It is recommended to set muffler in manual exhaust pipe. | {2, 3, 2} | Hengyi Petrochemical (http://www.hengyi.com/) |
| 100 thousand T / a sulfur recovery | Similar to the last process. | Acid gas pretreatment unit; Steam generator and superheater; Sulfur generation gas conversion, heat exchange and cooling unit; Tail gas treatment system; Sulfur forming system; Boiler blowdown system. | Because the clean acid gas carries too much solvent regeneration liquid, the liquid level of its knockout tank is too high, and the liquid level of tank V-1001 rises, which makes the liquid enter the combustion furnace F-1001, damages the combustion furnace equipment and bursts the furnace body. Add LI10101 high level alarm. It is recommended to consider V-1001 high level interlock. | {3, 2, 2} | Sichuan Petrochemical (http://scsh.cnpc.com.cn/) |
| 300 T / h solvent regeneration | The main element is solvent, and its heat transfer is flash evaporation at medium temperature. The solvent preparation can prevent the oxidation and deterioration of solvent by setting a water-sealed tank and shunt control. Solvent regeneration covers pressure and liquid level control of overhead reflux drum, steam flow control of bottom reboilers and bottom liquid level control. De-solvent is used to treat sulfur-containing gas. | Lean and rich liquid ring heating and its flash system; Solvent regeneration tower and condensate tank system; Top cooling system of solvent regeneration tower; Solvent preparation and recovery system; Amine liquid online purification system; Solvent storage tank system. | As the liquid carries gas into the gas flare header, the liquid level in the rich liquid flash tank is too low, which makes the safety valve PSV-5001 trip. Execute LIC50201, PIC50202, PSV5001 and PIA50201. It is recommended to recalculate the tripping pressure of safety valve PSV-5001. | {2, 5, 2} | Sichuan Petrochemical (http://scsh.cnpc.com.cn/) |
| 1 million T / a hydrocracking | Under high pressure and high temperature with hydrogen as catalyst, heavy oil is refined through hydrogenation, cracking and isomerization, and converted into light oil, i.e. gasoline, kerosene, diesel oil or raw materials for catalytic cracking and cracking to olefins. | Feed oil buffer tank and filter; Reaction water injection system; Separator system; Liquid separation system; Cooling system; Heat exchanger system; Return system; Feeding system; Hydrogenation system. | As the filter at the inlet of hydrogenation feed pump P-8102 is blocked, the liquid level of feed oil buffer tank V-8101 is too high, which causes the evacuation damage of P-8102 in serious cases. Prepare pump LICA1001A and high alarm LG1001. It is recommended to add a differential pressure gauge before and after the inlet filter screen. | {2, 3, 2} | Liaoyang Petrochemical (http://lysh.cnpc.com.cn/) |
| 350 thousand T / a polyethylene production | The catalyst and propane diluent are mixed into the pre-polymerization reactor, and the cocatalyst, ethylene, comonomer and hydrogen are simultaneously sent to. The pre-polymerized slurry enters the slurry loop reactor and operates under supercritical conditions to produce bimodal products. The polymer after flash evaporation is further sent to the fluidized bed gas phase reactor to obtain homopolymer. | Wastewater treatment and recovery system; Steam distillation system; Steam condensation recovery and circulation system; Nitrogen cooling and washing system; Powder homogenizing, drying, feeding and feeding systems; Unloading station, irrigation area and system for catalyst preparation, mixing and mixing; The system of reactant polymerization, centrifugation and filtration. | Due to PIC14201~14204 fault, the flow rate of steam entering 11403a/b is too low, the separation effect is poor, the content of diluent is high, and the volatile is ignited and exploded when encountering the fire source. Hold 11504a alarm and interlock motor protection and interlock. It is recommended to introduce the distillation kettle pressure PI14433 into DCS. | {4, 2, 2} | Liaoyang Petrochemical (http://lysh.cnpc.com.cn/) |
| 8 thousand T / a cis polybutadiene rubber production | The prepared and aged catalyst, monomer butadiene and solvent are pumped into the polymerization unit to synthesize cis-polybutadiene rubber. The glue shall be added with terminator and anti-aging agent before setting, and further dehydrated and dried after being condensed by steam. The remaining solvent oil and butadiene are refined and recycled. | Preparation tank system; High level tank system; Polymerization system; Pre-mixing kettle system; Storage system; Stripping kettle system; Buffer system; Recovery system; Waste gas treatment system; Heat exchanger system; Absorber system; Steam system; Flare system; Discharge system. | Due to the high concentration of butadiene, the temperature of the first polymerization kettle is too high, so that the reaction temperature is not out of control and the polymerization is explosive when it is serious, the agitator and kettle equipment are damaged, the materials in the kettle enter the low-pressure pipe network, and the pipeline is blocked. TRCA11014 high alarm, R-1101 over temperature interlock, TRA11017 high alarm, TRA11018 high alarm and R-1101 over pressure interlock are added. It is suggested that $R-1101$ overpressure interlocking system should be set with independent detection points in the design. | {2, 5, 3} | Sichuan Petrochemical (http://scsh.cnpc.com.cn/) |
| 200 T / h acid water stripping | The acid water from the atmospheric and vacuum distillation unit and the heavy oil catalytic cracking unit is sent to the acid water degassing tank, and the light hydrocarbon components are dialyzed and sent to the low-pressure gas pipe network. The remaining materials enter the sour water tank for oil removal, standing and sedimentation. The upper waste oil is collected and sent to the waste oil tank for treatment. | Acid water collection, degassing and storage system; Sour water stripper and its purification system; Deodorization and its absorption tower system; Circulating water system; Feed condensation, desalination, filtration, purification and steam system. | Due to the large amount of acid water, the pressure of the acid water degassing tank is too high, so that the liquid level and pressure of the tank rise. In serious cases, the gas carries liquid into the flare header. Prepare LIC30101 high alarm and PI30101 high alarm. It is recommended that the pipeline from the tank top to the main gas flare vent pipe be designed without liquid. | {2, 5, 2} | Sichuan Petrochemical (http://scsh.cnpc.com.cn/) |
| 500 thousand T / a gas fractionation | The liquid hydrocarbon after desulfurization and mercaptan removal enters the depropanizer. Liquefied petroleum gas is distilled to distillate at the tower top. One part of the condensate is sent to the depropanizer for reflux, and the other part is sent to the deethanizer as feed. The bottom material fraction of depropanizer is sent out of the unit after being cooled by C4/C5 cooler. | Depropanizer and its feed and reflux system; Deethanizer system; Propylene rectifying tower, debutanizer and their reflux systems; Gas purification and compression system; Steam system; Circulating water system. | Due to FRC10105 circuit fault, the opening of FV10105 is large, which makes the temperature of depropanizer too high, causing the temperature of tower C-1001 to rise, the pressure to rise, the liquid level to drop, and the tower to be flushed. Add PRC10101 and LICA10102 low alarm. It is recommended to design TII10101 to TII10105 high alarm. | {2, 4, 1} | Sichuan Petrochemical (http://scsh.cnpc.com.cn/) |

It should be noted that these enterprises are representative, and their HAZOP reports are standard and high-quality. Different processes and enterprises make HAZOP reports different, and there is no comparability between them in this study. Another reminder is that the HAZOP report is protected by intellectual property rights and confidential.

## 5. EXPERIMENT & ANALYSIS

### 5.1. Datasets

Table 3: Information on HaE datasets.

| Aspect | Level #1 | Level #2 | Level #3 | Level #4 | Level #5 |
|---|---|---|---|---|---|
| Possibility | 419 | 1760 | 1607 | 1134 | 949 |
| Severity | 1570 | 2732 | 1353 | 170 | 44 |
| Risk | 2902 | 2577 | 335 | 55 | - |

HaE from research cases can ensure the universality of experiments and evaluate our methods well. Through text preprocessing, we collect 5869 HaE with labels from research cases as datasets for each classification aspect, see Table 3 for the quantities of different levels. We randomly assign them into training set, test set and validation set at a ratio of 8:1:1.

*5.2. Trial model*

The models used for comparative experiments are:

1. BERT (base#1): The feature vectors generated by BERT are directly transmitted to a fully connected neural network (FC) for classification prediction (Feng et al., 2021).

2. BERT-CNN (base#2): The feature vector encoded by BERT is further extracted by CNN, and then decoded by an FC to predict classification (Lu et al., 2022).

3. BERT-RAtt (base#3): BiLSTM and Attention jointly act as the encoder for feature vectors operated by BERT in advance, and an FC completes classification prediction (Feng et al., 2021).

4. BERT-DPCNN (base#4): In brief, the feature vectors generated by BERT are encoded by DPCNN that is a network composed of multiple isometric-convolution layers and *half*-pooling operations, and then an FC implements decoding classification (Li and Ning, 2020).

5. BERT-RCNN (base#5): In short, benefiting from computer vision, BiLSTM and CNN with *max*-pooling jointly act as the encoder for feature vectors, and an FC completes classification prediction (Qin et al., 2021).

There are two modules in DLF that should be evaluated: HmF-DFA and HGNN. The models used for ablation experiments are:

6. BERT-HGNN (DLF#1).

7. BERT-[HmF-DFA] (DLF#2): An FC is used for the final classification prediction.

8. BERT-[mF-DFA] (DLF#3): An FC is used for the final classification prediction.

9. BERT-[mF-DFA]-HGNN (DLF#4).

10. BERT-[HmF-DFA]-HGNN (DLF#5).

Where, DLF#5 is our complete model; DLF#1 *vs.* DLF#5 for evaluating the benefit of HmF-DFA; DLF#2 *vs.* DLF#3 and DLF#4 *vs.* DLF#5 for evaluating the progress of HmF-DFA compared with mF-DFA; DLF#2 *vs.* DLF#5 and DLF#3 *vs.* DLF#4 for evaluating the effect of HGNN.

*5.3. Experiment setting*

In each trial experiment, the main parameters are consistent. For example, the size of BERT is base, the optimizer is Adam with a learning rate of 1e-5, the epoch of training is 50, and the batch size is 128. We take the average results of 5 repetitions as the evaluation report. The evaluation metrics are F1-score (F1), precision (P) and recall (R) (Feng et al., 2021), see Equ.22 and Table 4. Where, P means the proportion of actual positive samples among all predicted positive samples, R means the proportion of predicted positive samples among the actual positive samples, F1 considers the balance of P and R, which is their harmonic average.

$$P = \frac{TP}{TP + FP}, \quad R = \frac{TP}{TP + FN}, \quad F1 = \frac{2PR}{P + R} \qquad (22)$$

Table 4: Information on evaluation metrics.

| Predicted \ Actual | Positive | Negative |
|---|---|---|
| True | True Positive (TP) | False Negative (FN) |
| False | False Positive (FP) | True Negative (TN) |

*5.4. Evaluation analysis*

Table 5-7 present all the total classification results of different models under the three aspects respectively, where, "test" means the test set and "val" means the validation set. There are some major observations.

Obviously, the performance of our DLF#5 model has an overwhelming superiority over that of the base series models, especially the severity aspect. In addition, for the F1, our model leads other models under all three aspects both on the test set and validation set. See Fig.2-4 for a more intuitive presentation, where the ordinate represents a pair of models, and the abscissa represents the performance gap between them. For example, "DLF#5 / base#1" represents how much higher the performance of DLF#5 is than that of base#1. It can be seen that the performance of our model on the severity aspect is significantly higher than that of previous models, especially compared with base#3, base#4 and base#5, most of the leading ranges even exceed six percentage points. On the possibility aspect, our model has slight weakness only in two evaluations, and its performance in the remaining 28 evaluations is remarkable. The same is true of the performance under the risk aspect, with only one exception. Undoubtedly, our DLF classifier is advanced and gratifying, and it can serve HaE's classification work more effectively.

Table 5: Evaluation results (%) under the severity aspect.

| Model | P | | R | | F1 | |
|---|---|---|---|---|---|---|
| | test | val | test | val | test | val |
| BERT (base#1) | 81.23 | 78.09 | 85.54 | 82.17 | 82.92 | 79.77 |
| BERT-CNN (base#2) | 78.46 | 78.06 | 82.04 | 81.69 | 79.30 | 79.31 |
| BERT-RAtt (base#3) | 77.86 | 75.58 | 83.93 | 78.30 | 79.99 | 76.11 |
| BERT-DPCNN (base#4) | 76.03 | 73.91 | 77.92 | 76.04 | 75.66 | 74.67 |
| BERT-RCNN (base#5) | 75.82 | 74.91 | 81.78 | 79.69 | 77.06 | 76.24 |
| BERT-HGNN (DLF#1) | 81.12 | 78.59 | 85.57 | 82.62 | 82.99 | 80.25 |
| BERT-[HmF-DFA] (DLF#2) | 81.55 | 80.67 | 84.99 | 84.86 | 83.06 | 82.33 |
| BERT-[mF-DFA] (DLF#3) | 80.84 | 80.87 | 84.27 | 80.88 | 81.86 | 80.30 |
| BERT-[mF-DFA]-HGNN (DLF#4) | 82.46 | 81.95 | **86.36** | **85.70** | 83.85 | 83.37 |
| BERT-[HmF-DFA]-HGNN (DLF#5) | **83.04** | **82.75** | 85.99 | 85.13 | **84.09** | **83.38** |

Table 6: Evaluation results (%) under the possibility aspect.

| Model | P | | R | | F1 | |
|---|---|---|---|---|---|---|
| | test | val | test | val | test | val |
| BERT (base#1) | 69.97 | 69.14 | 70.73 | 70.20 | 70.18 | 69.63 |
| BERT-CNN (base#2) | 69.44 | 69.22 | 72.54 | 70.29 | 70.73 | 69.70 |
| BERT-RAtt (base#3) | 69.12 | 68.96 | 71.81 | 72.03 | 70.11 | 70.08 |
| BERT-DPCNN (base#4) | 69.71 | 68.62 | 72.51 | 72.35 | 70.20 | 69.98 |
| BERT-RCNN (base#5) | **70.62** | 69.31 | 71.90 | 72.12 | 70.95 | 70.16 |
| BERT-HGNN (DLF#1) | 69.66 | 69.69 | 72.56 | 73.19 | 70.61 | 70.85 |
| BERT-[HmF-DFA] (DLF#2) | 68.90 | 69.88 | **74.31** | **73.88** | 70.56 | 71.32 |
| BERT-[mF-DFA] (DLF#3) | 70.38 | 70.25 | 69.89 | 70.84 | 70.04 | 70.48 |
| BERT-[mF-DFA]-HGNN (DLF#4) | 68.59 | 70.16 | 72.52 | 73.22 | 70.12 | 71.25 |
| BERT-[HmF-DFA]-HGNN (DLF#5) | 70.45 | **70.41** | 72.52 | 73.29 | **71.35** | **71.51** |

Table 7: Evaluation results (%) under the risk aspect.

| Model | P | | R | | F1 | |
|---|---|---|---|---|---|---|
| | test | val | test | val | test | val |
| BERT (base#1) | 69.48 | 71.51 | 74.33 | 71.97 | 71.30 | 71.73 |
| BERT-CNN (base#2) | 70.52 | **73.68** | 71.66 | 70.68 | 71.05 | 71.92 |
| BERT-RAtt (base#3) | 70.92 | 69.23 | 70.59 | 73.33 | 70.75 | 70.99 |
| BERT-DPCNN (base#4) | 67.34 | 69.97 | 76.75 | 70.19 | 70.84 | 69.97 |
| BERT-RCNN (base#5) | 68.59 | 69.80 | 74.00 | 70.48 | 70.93 | 70.13 |
| BERT-HGNN (DLF#1) | **74.45** | 73.09 | 74.33 | 75.17 | 74.30 | 74.02 |
| BERT-[HmF-DFA] (DLF#2) | 69.44 | 72.54 | 78.38 | 74.17 | 72.95 | 73.22 |
| BERT-[mF-DFA] (DLF#3) | 72.50 | 73.10 | 74.98 | 72.80 | 73.27 | 72.93 |
| BERT-[mF-DFA]-HGNN (DLF#4) | 70.60 | 72.39 | **79.54** | 74.63 | 74.11 | 73.45 |
| BERT-[HmF-DFA]-HGNN (DLF#5) | 73.07 | 71.44 | 77.15 | **79.02** | **74.77** | **74.45** |

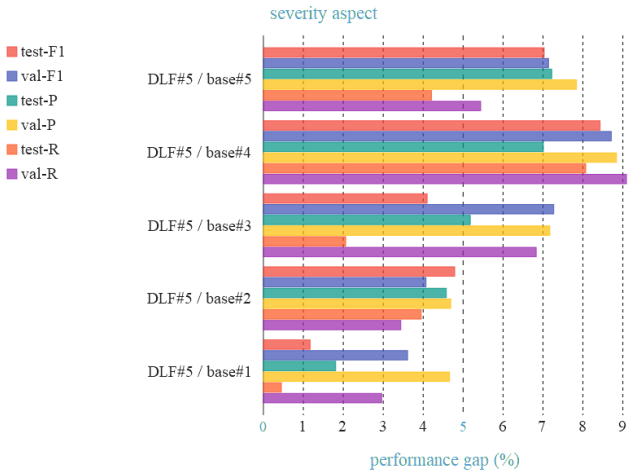

Fig.2: The performance gap between DLF#5 and the base series models under the severity aspect.

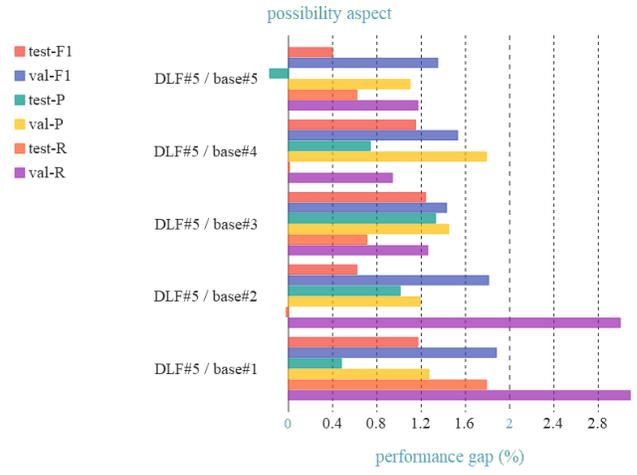

Fig.3: The performance gap between DLF#5 and the base series models under the possibility aspect.

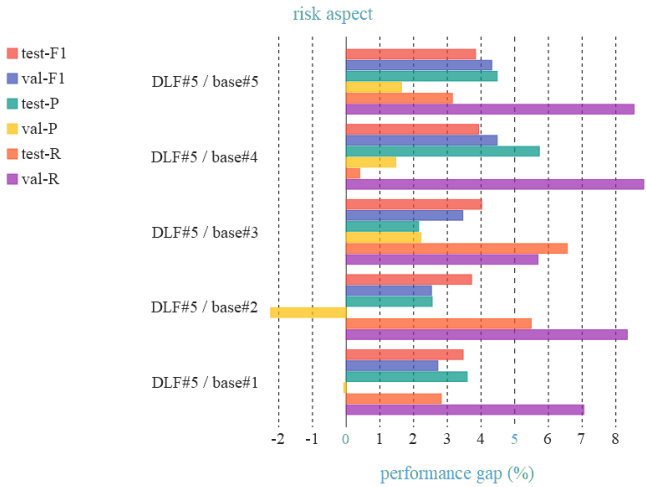

Fig.4: The performance gap between DLF#5 and the base series models under the risk aspect.

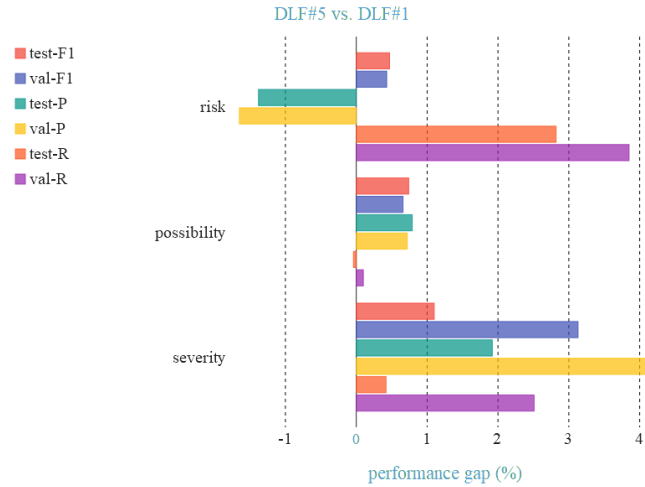

Fig.5: DLF#5 *vs*. DLF#1 for evaluating the profit of HmF-DFA.

We evaluate the profit of the proposed HmF-DFA through the performance gap between DLF#5 and DLF#1, see Fig.5. It can be seen that in three aspects, except for three exceptions, all the evaluations reflect that HmF-DFA has a positive promotion on the DLF classifier. Therefore, HmF-DFA is feasible and effective for HaE's classification duty.

We employ the performance gap between DLF#2 and DLF#3, as well as, that between DLF#5 and DLF#4, to evaluate the progress of HmF-DFA compared with mF-DFA. See Fig.6-7, it can be observed that although a small number of evaluations reflect retrogression, most of them are weak. While other evaluations show that HmF-DFA tends to be more progressive, such as the evaluations under the severity aspect in Fig.6, and that under the possibility aspect in Fig.7. Therefore, HmF-DFA is more suitable than mF-DFA and more conducive to the classification responsibilities of HaE.

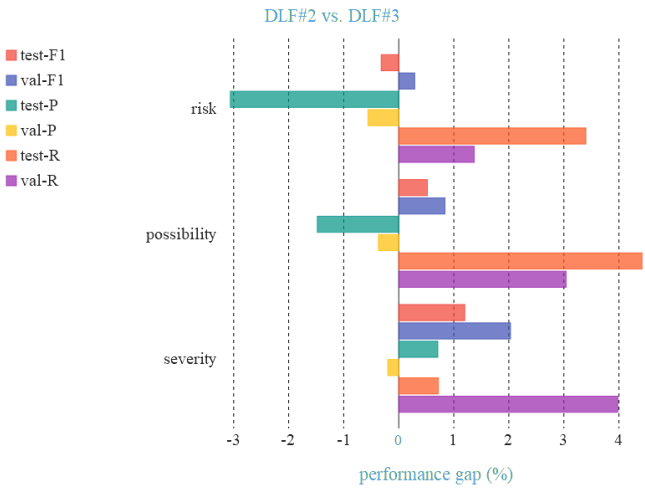

Fig.6: DLF#2 *vs.* DLF#3 for evaluating the progress of HmF-DFA.

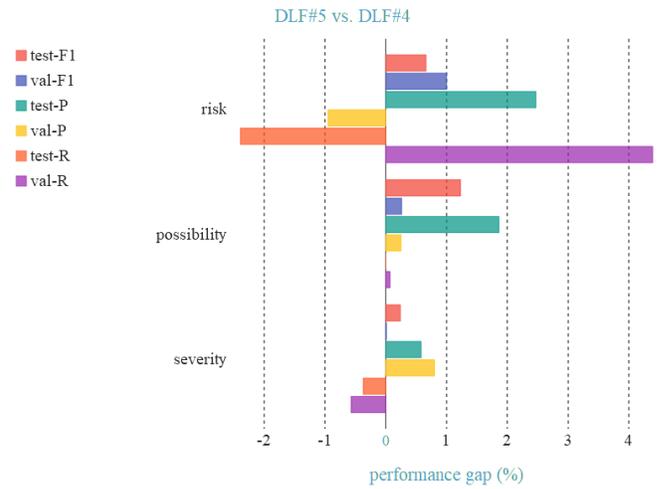

Fig.7: DLF#5 *vs.* DLF#4 for evaluating the progress of HmF-DFA.

DLF#5 *vs.* DLF#2 and DLF#4 *vs.* DLF#3 are used to measure the gain of HGNN, see Fig.8-9. It can be clearly seen that the vast majority of evaluations indicate that HGNN has indeed brought considerable performance gains to the DLF classifier, and it has even improved by more than four percentage points in some evaluations, especially under the severity aspect, and all the evaluations are friendly. Undoubtedly, HGNN has greatly enhanced the classification of HaE.

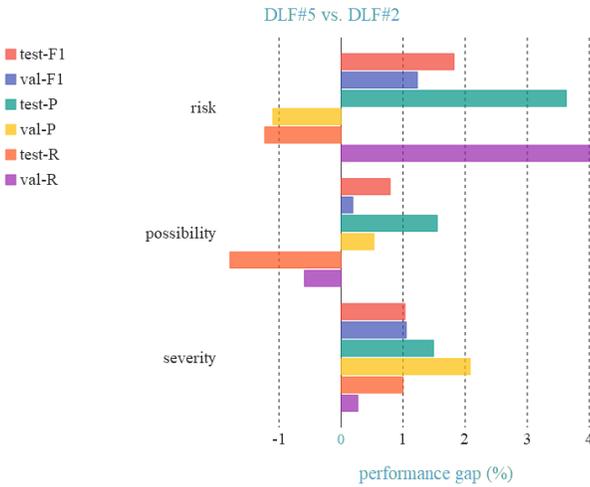

Fig.8: DLF#5 vs. DLF#2 for evaluating the gain of HGNN.

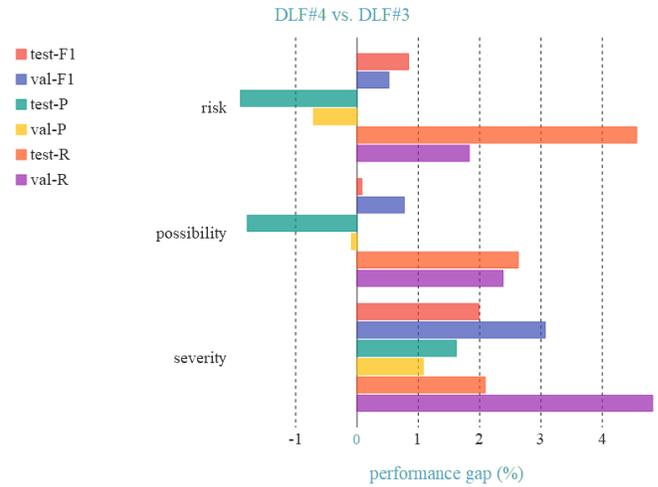

Fig.9: DLF#4 vs. DLF#3 for evaluating the gain of HGNN.

To sum up, our DLF classifier has promising and gratifying aptitudes, does ameliorates the classification work of HaE. We explore HaE based on deep learning from the perspective of language through the concept of fractal, which is novel and profound. We hope our research can contribute added value to the daily practice in industrial safety and provide support for fractal theory.

## 6. APPLICATION

We embed the above trained DLF classifier as the HaE classification system into industrial safety knowledge graph (ISKG) (Wang et al., 2022) to serve and support HAZOP, see Fig.10. At present, HaE classification system can mainly undertake the following preliminary auxiliary applications.

### 6.1. For the expert group

Assist the expert group to conduct safety analysis on the raw process and support the identification of decision-making. There are large-scale nodes and their intricate objective connections and causalities in the process, hence HAZOP needs to consume a lot of manpower and time in confront new processes, and its analysis efficiency is inevitably flawed. Relievedly, HaE classification system can be used as an auxiliary proofreading to alleviate this embarrassment. Experts can flexibly check and review the hazard that the recorded level is inconsistent with the inference given by HaE classification system during the calibration of analysis results, so as to reduce work costs and potential mistakes caused by human boundedness.

Take two pre-analyzed HaE, called HaE#1 and HaE#2 respectively, in 2.2 million T / a diesel hydrofining process and 4 million T / a indirect coal liquefaction process in Table 2 as cases. Where, they exist on the R-5611101 (a kind of Fischer Tropsch reactor), and the level of {severity / possibility / risk} given by experts to them in advance are {5 / 5 / 2} and {3 / 4 / 2} respectively.

During the review, HaE classification system first gives its own predictions for these two, then compares the prediction results with the inferences of the experts (can draw support from ISKG), and finally, (1) for the hazards with inconsistent comparison results, they are fed back to the experts in the order of the scale of level of severity, possibility and risk given by the experts for review in turn. For instance, the predictions given to HaE#1 and HaE#2 are {4 / 3 / 3} and {4 / 2 / 1} respectively, which is inconsistent with the inference of the expert group, hence, the two are reviewed in priority, and the priority of the former is higher than that of the latter. (2) For the HaE with consistent comparison results, they are placed in the final unified review, and the order in procedure (1) is also followed. Where, the rationality of the suggestions and the suitability of measures need more rigorous proofreading.

In addition, in the subsequent development of this application, it can also appropriately guide the expert group to make the analysis during the pre-brainstorming to reduce the workload.

Therefore, our HaE classification system can support the expert group to complete the safety exploration for new processes.

### 6.2. For the engineer

Assist the engineer in dealing with unforeseen hazards. For the process that has been put into production, there are inevitably additional hazards that have not been analyzed by the expert group. Fortunately, engineers can draw support from HaE classification system to qualitatively analyze such hazards in advance, and then quickly promote appropriate solutions and aftercare management according to the predicted attributes {severity / possibility / risk} of the hazards. Suppose such a hazard ignored by experts is triggered: a local explosion in the natural gas hydrogen production process. The engineer traces the source of this hazard (can work with ISKG):

E0102A/B heat exchanger shakes electricity, which makes E0103 corroded, and the converted gas leaks into the shell side of E0103, resulting in the explosion of shell side process after overpressure leakage.

At this time, HaE classification system can immediately give the attribute {3 / 3 / 2} of the hazard and confirms or modifies it with the understanding of the engineer. After the attribute is approved, HaE classification system implements (also approved by the engineer) the solutions that meet or are similar to the attribute {3 / 3 / 2} retrieved from the E0103 related HAZOP library of natural gas hydrogen production process, so as to mitigate the spread and secondary injury of the hazard as quickly as possible, for example, set the normally open vent pipeline and PRV0155A on the deaerator. Meanwhile, relevant suggestions are fed back to the engineer, for example, regularly test the hydrogen concentration at the exhaust outlet on the top of the deaerator and monitor the leakage of E0103.

Undoubtedly, HaE classification system can help engineers to quickly combat unforeseen hazards.

### 6.3. For the employee

Support the employee to conduct routine work and emergency response. In the daily workflow such as the system maintenance and troubleshooting, with the help of the reminder of HaE classification system, the employee can conveniently adjust the check scheduling according to the type of hazards. Specifically, HaE classification system acts as a medium of information communication, and employees can give priority to eliminating hazards with high possibility level and pay more attention to hazards with strong severity level according to the provided information about potential hazards involved, etc.

Moreover, in terms of emergency response, it is common for a single deviation factor to cause multiple hazards to be triggered at the same time. Gratifyingly, HaE classification system can participate in the security scheduling of the process, it can sort the urgency of hazards according to the size of their severity level, so as to appropriately and orderly assist employees in making decisions about scheduling on how to deal with multiple hazards.

Accordingly, HaE classification system can facilitate employees to complete their work.

### 6.4. For other related enterprises

Guide related processes of other enterprises to launch HAZOP. For some small-scale enterprises and independent processes, the received safety analysis is often not exhaustive, since the quality of HAZOP and completeness of HaE are subject to expert teams and processes of different scales. Specifically, for the Claus reactor, the working conditions it faces in different processes may be different, and it interacts with other equipment and materials to trigger different hazards whose diversity and complexity increase with the expansion of the scale of the process. Hence, the Claus reactor is often more likely to be analyzed more thoroughly in large-scale processes, rather than the small-scale processes. In addition, the analysis power of expert teams in different enterprises is also different, which is related to the empirical competence and knowledge reserve of experts. Some small-scale enterprises feebly analyze and collect the hazards related to the Claus reactor as comprehensively as possible.

Therefore, it is necessary for small-scale enterprises to enhance the safety of related processes through the guidance of high-quality HAZOP knowledge (Feng et al., 2021). However, HAZOP report is confidential, protected by property rights, and is a scarce resource.

Hearteningly, our HaE classification system absorbs the HAZOP knowledge of a number of large-scale enterprises involving a total of 18 processes, and is competent for this guidance. Its working mode is similar to the first three applications, which will not be repeated here. To a certain extent, our work has brought progress to the transfer and sharing of knowledge.

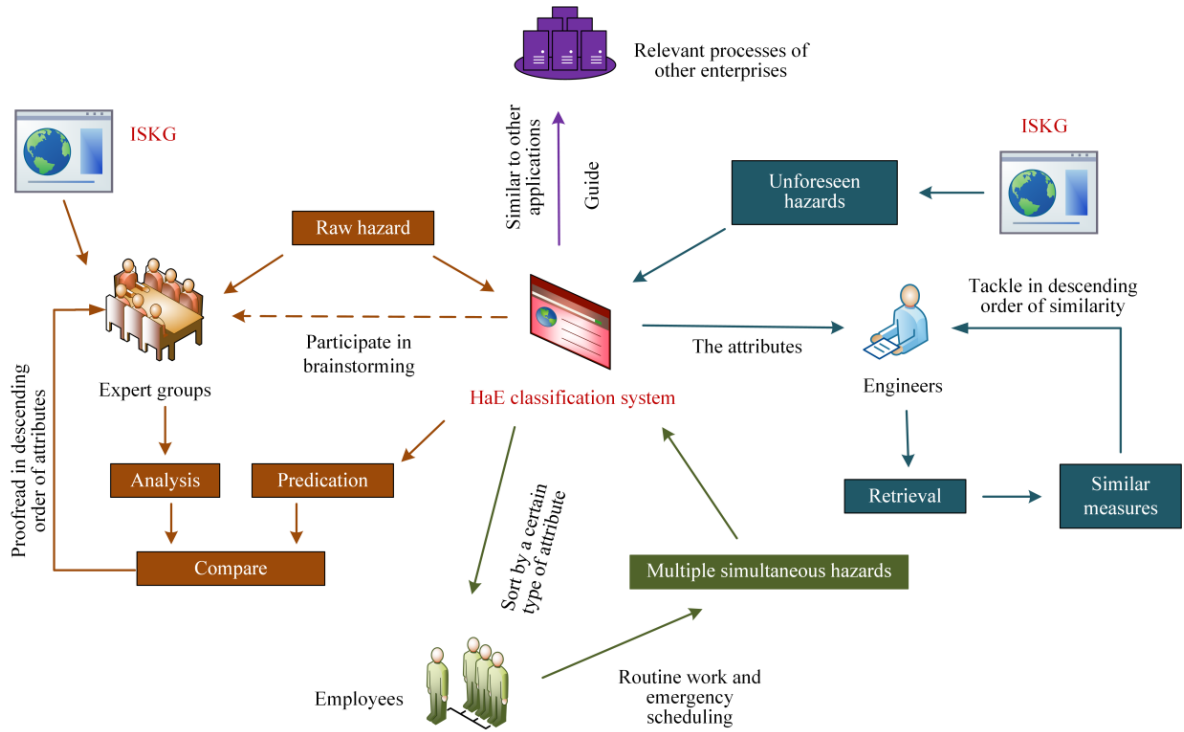

Fig.10: Application practice of the HaE classification system, where, the attributes refer to {severity, possibility, risk} of the HaE.

## 7. DISCUSSION

The proposed DLF model is driven by deep learning and inspired by multifractal. Its excitement can be attributed to the fusion of local features and context features, as well as the additional profound stimulation brought by language fractal features. One idea is that HaE is naturally regarded as a kind of time series, and has an opportunity to be explored through multifractal methods. On this basis, the DLF trained by 18 processes undertakes the HaE classification system, which has practical value, can bring application incentives to experts, engineers, employees, and other enterprises.

It should be noted that there are some limitations that need to be clarified.

From the perspective of DLF model, No Free Lunch Theorem (Ho and Pepyne, 2002) teaches us that no model can face all practical issues. Similarly, DLF model also faces the limitation of how it can be migrated to other classification tasks. The DLF model may need to be adjusted to adapt to different domains. For example, some hazard events may be incomplete due to the quality of technical reports, and DLF model may be less robust. In addition, DLF model is suitable for hazards with certain time series attributes, while it may be powerless for ordinary hazards. Moreover, the sample imbalance limits the performance of the DLF model. Although HmF-DFA considers this, it is not enough, since with the participation of different types of processes, the sample imbalance becomes more serious. We hope that the follow-up work needs further consideration, such as data enhancement.

From the point of view of HaE classification system, for the availability of public use, one limitation is to provide reference for the safety analysis of other types of processes, because our existing HaE classification system is mainly based on the petrochemical industry, and weakly serve and communicate with other types of industries, such as aviation related industries and civil engineering related ones. Hence, we need to strengthen and harmonize the audience size and popularity of HaE classification system. In addition,

HaE classification system can indeed deliver value with ISKG, but it is affected by the quality of ISKG. For example, if the unexpected HaE inferred by ISKG to engineers are wrong, HaE classification system will fail. In this case, the classification system alone is feebly affordable, which is also a boundary to be optimized. Another limitation is that although the application of HaE classification system benefits from HaE with its measures and suggestions, it is inevitable that some HaE have no suggestions, which will affect the predicted results.

Sincerely, we hope that our research can burst out its due value and serve the intelligent progress of industrial safety more comprehensively and orderly.

## 8. CONCLUSION

The national volume and industrial development have made HAZOP a leader in industrial safety engineering. Therefore, it is meaningful and necessary to study its HaE classification, which has far-reaching significance for the intelligent development and progress of industrial safety. In view of this, this paper proposes a new deep learning-based classifier DLF to explore HaE through multifractal method from three aspect: severity, possibility and risk. Numerous experiments prove the effectiveness and progressiveness of DLF, which depends on the fractal processing of HmF-DFA, as well as the feature fusion of HGNN. Where, the former is based on the fact that HaE is regarded as a kind of time series and are vectorized by BERT. The latter skillfully controls CNN and BiLSTM. HaE classification system is beneficial to the redevelopment and enhancement of HAZOP, it can serve the safety and reliability of industry intelligently and orderly by promoting the expert group to conduct safety analysis, assisting the engineer to treat unforeseen hazards, supporting employees to conduct routine work, and guiding related processes of other enterprises to launch HAZOP. Our research is promising and meaningful, we hope that it can bring encouragement and inspiration to other researchers.

### REFERENCE

Andres, J., Langer, J., & Matlach, V. (2020). Fractal–based analysis of sign language. Communications in Nonlinear Science and Numerical Simulation, 84, 105214.

Barber, S. , Boulinquiez, M. , Caprio, E. D. , Candeal, J. , & Kernen, H. . (2017). Hazard classification and labelling of petroleum substances in the european economic area - 2017. CONCAWE Reports(13), 1-317.

Cheraghi, M. , Baladeh, A. E. , & Khakzad, N. . (2019). A fuzzy multi-attribute HAZOP technique (FMA-HAZOP): application to gas wellhead facilities. Safety Science, 114, 12-22.

China General Administration of work safety. (July 2013). Guidance on strengthening safety management of chemical process, 31. www.gov.cn/gongbao/content/2013/content_25197 13.htm.

Choubin, B., Mosavi, A., Alamdarloo, E. H., Hosseini, F. S., Shamshirband, S., Dashtekian, K., & Ghamisi, P. (2019). Earth fissure hazard prediction using machine learning models. Environmental research, 179, 108770.

Clemente, G. P., & Cornaro, A. (2022). A multilayer approach for systemic risk in the insurance sector. Chaos, Solitons & Fractals, 162, 112398.

Deng, J., Cheng, L., & Wang, Z. (2021). Attention-based BiLSTM fused CNN with gating mechanism model for Chinese long text classification. Computer Speech & Language, 68, 101182.

Emergency Management Department of the people's Republic of China. (July 2019). Management measures for emergency management standardization. https://www.mem.gov.cn/gk/tzgg/ tz/201907/t20190707_321229.shtml.

Emergency Management Department of the people's Republic of China. (November 2020). Catalogue of safety classification and rectification of hazardous chemical enterprises. https://www.mem.gov.cn/gk/tzgg/tz/202011/t20201103_371291.shtml. Accessed 2022.

Emergency Management Department of the people's Republic of China. HAZOP. https://www.mem.gov.cn/. 2022.

Fang, W., Luo, H., Xu, S., Love, P. E., Lu, Z., & Ye, C. (2020). Automated text classification of near-misses from safety reports: An improved deep learning approach. Advanced Engineering Informatics, 44, 101060.

Fattor, M. V. , & Vieira, M. A. . (2019). Application of human HAZOP technique adapted to identify risks in brazilian waste pickers' cooperatives. Journal of Environmental Management, 246(SEP.15), 247-258.

Feng, X., Dai, Y., Ji, X., Zhou, L., & Dang, Y. (2021). Application of natural language processing in HAZOP reports. Process Safety and Environmental Protection, 155, 41-48.

Ferreira, C. , Ribeiro, J. , & Freire, F. . (2019). A hazard classification system based on incorporation of reach regulation thresholds in the usetox method. Journal of Cleaner Production, 228(AUG.10), 856-866.

Gaurav, G., Anand, R. S., & Kumar, V. (2021). EEG based cognitive task classification using multifractal detrended fluctuation analysis. Cognitive Neurodynamics, 15(6), 999-1013.

Gómez-Gómez, J., Carmona-Cabezas, R., Ariza-Villaverde, A. B., de Ravé, E. G., & Jiménez-Hornero, F. J. (2021). Multifractal detrended fluctuation analysis of temperature in Spain (1960–2019). Physica A: Statistical Mechanics and its Applications, 578, 126118.

Gu, A., Gulcehre, C., Paine, T., Hoffman, M., & Pascanu, R. (2020, November). Improving the gating mechanism of recurrent neural networks. In International Conference on Machine Learning (pp. 3800-3809). PMLR.

Gui, J., Zheng, Z., Fu, D., Fu, Y., & Liu, Z. (2021). Long-term correlations and multifractality of toll-free calls in China. Physica A: Statistical Mechanics and its Applications, 567, 125633.

Hou, R., Huang, C. R., Do, H. S., & Liu, H. (2017). A study on correlation between Chinese sentence and constituting clauses based on the Menzerath-Altmann law. Journal of Quantitative Linguistics, 24(4), 350-366.

Ho, Y. C., & Pepyne, D. L. (2002). Simple explanation of the no-free-lunch theorem and its implications. Journal of optimization theory and applications, 115(3), 549-570.

Hrjs, A. , Sn, B. , Ah, A. , Hm, C. , & Mk, D. (2022). A fuzzy-HAZOP/ant colony system methodology to identify combined fire, explosion, and toxic release risk in the process industries. Expert Systems with Applications, 192, 116418.

Ishola, A. A., & Valles, D. (2022, June). Using Machine Learning and Regression Analysis to Classify and Predict Danger Levels in Burning Sites. In 2022 IEEE World AI IoT Congress (AIIoT) (pp. 453-459). IEEE.

Jahani, A. (2019). Sycamore failure hazard classification model (SFHCM): an environmental decision support system (EDSS) in urban green spaces. International journal of environmental science and technology, 16(2), 955-964.

Jeon, J., & Cai, H. (2021). Classification of construction hazard-related perceptions using: Wearable electroencephalogram and virtual reality. Automation in Construction, 132, 103975.

Jing, S., Liu, X., Gong, X., Tang, Y., Xiong, G., Liu, S., ... & Bi, R. (2022). Correlation analysis and text classification of chemical accident cases based on word embedding. Process Safety and Environmental Protection, 158, 698-710.

Júnior, J. S., Paulo, J. R., Mendes, J., Alves, D., Ribeiro, L. M., & Viegas, C. (2022). Automatic forest fire danger rating calibration: Exploring clustering techniques for regionally customizable fire danger classification. Expert Systems with Applications, 193, 116380.

Khan, M. K., Imran, M., Ahmad, M. H., Ahmad, R. S., Chemat, F., & Zulifqar, A. (2022). Food Safety and Security (HACCP and HAZOP) for Consumers and Workers (Nonthermal Technologies and Their Use). In Nonthermal Processing in Agri-Food-Bio Sciences (pp. 749-768). Springer, Cham.

Khan, N., Saleem, M. R., Lee, D., Park, M. W., & Park, C. (2021). Utilizing safety rule correlation for mobile scaffolds monitoring leveraging deep convolution neural networks. Computers in Industry, 129, 103448.

Kletz, T. (2018). HAZOP and HAZAN: identifying and assessing process industry hazards. CRC Press.

Kubota, M., & Zouridakis, G. (2022). Differentiation of task complexity in long-term memory retrieval using multifractal detrended fluctuation analysis of fNIRS recordings. Experimental Brain Research, 1-11.

Lancaster, L., Ostriker, E. C., Kim, J. G., & Kim, C. G. (2021). Efficiently Cooled Stellar Wind Bubbles in Turbulent Clouds. I. Fractal Theory and Application to Star-forming Clouds. The Astrophysical Journal, 914(2), 89.

Li, J., Sun, Y., Wang, G., Liu, Y., & Sun, Z. (2022). Different discrete-time noise-suppression Z-type models for online solving time-varying and time-invariant cube roots in real and complex domains: Application to fractals. Neurocomputing. 500, 471-485.

Lim, C. H. , Lam, H. L. , & Pei Qin, W. N. . (2018). A novel HAZOP approach for literature review on biomass supply chain optimisation model. Energy. 146, 13-25.

Lim, C. H. , Lim, S. , Bing, S. H. , Ng, W. , & Lam, H. L. . (2021). A review of industry 4.0 revolution potential in a sustainable and renewable palm oil industry: HAZOP approach. Renewable and Sustainable Energy Reviews, 135, 110223.

Lin, J., Dou, C., & Liu, Y. (2021). Multifractal detrended fluctuation analysis based on optimized empirical mode decomposition for complex signal analysis. Nonlinear Dynamics, 103(3), 2461-2474.

Li, X., & Ning, H. (2020, September). Deep Pyramid Convolutional Neural Network Integrated with Self-attention Mechanism and Highway Network for Text Classification. In Journal of Physics: Conference Series (Vol. 1642, No. 1, p. 012008). IOP Publishing.

Li, Y., Zhu, Z., Kong, D., Han, H., & Zhao, Y. (2019). EA-LSTM: Evolutionary attention-based LSTM for time series prediction. Knowledge-Based Systems, 181, 104785.

Lu, H., Ehwerhuepha, L., & Rakovski, C. (2022). A comparative study on deep learning models for text classification of unstructured medical notes with various levels of class imbalance. BMC Medical Research Methodology, 22(1), 1-12.

Marhavilas, P. K. , Filippidis, M. , Koulinas, G. K. , & Koulouriotis, D. E. . (2021). Safety-assessment by hybridizing the mcdm/ahp & HAZOP-DMRA techniques through safety's level colored maps: implementation in a petrochemical industry-sciencedirect. Alexandria Engineering Journal. 61, 9, 6959-6977.

Meira, R. E., De Lai, F. C., Negrão, C. O., & Junqueira, S. L. (2020). On determining the power-law fluid friction factor in a partially porous channel using the lattice Boltzmann method. Physics of Fluids, 32(9), 093104.

Melluso, N., Grangel-González, I., & Fantoni, G. (2022). Enhancing Industry 4.0 standards interoperability via knowledge graphs with natural language processing. Computers in Industry, 140, 103676.

Mielniczuk, J., & Wojdyłło, P. (2007). Estimation of Hurst exponent revisited. Computational statistics & data analysis, 51(9), 4510-4525.

Miloş, L. R., Haţiegan, C., Miloş, M. C., Barna, F. M., & Boţoc, C. (2020). Multifractal detrended fluctuation analysis (MF-DFA) of stock market indexes. Empirical evidence from seven central and eastern European markets. Sustainability, 12(2), 535.

Minai, A. A., & Williams, R. D. (1993). On the derivatives of the sigmoid. Neural Networks, 6(6), 845-853.

Mwema, F. M., Akinlabi, E. T., Oladijo, O. P., Fatoba, O. S., Akinlabi, S. A., & Tălu, S. (2020). Advances in manufacturing analysis: Fractal theory in modern manufacturing. In Modern manufacturing processes (pp. 13-39). Woodhead Publishing.

Naeem, M. A., Farid, S., Ferrer, R., & Shahzad, S. J. H. (2021). Comparative efficiency of green and conventional bonds pre-and during COVID-19: An asymmetric multifractal detrended fluctuation analysis. Energy Policy, 153, 112285.

Najafi, E., Valizadeh, A., & Darooneh, A. H. (2022). The Effect of Translation on Text Coherence: A Quantitative Study. Journal of Quantitative Linguistics, 29(2), 151-164.

Nguyen, T. T., Hoffmann, E., & Buerkert, A. (2022). Spatial patterns of urbanising landscapes in the North Indian Punjab show features predicted by fractal theory. Scientific reports, 12(1), 1-14.

Ouyang, L., Che, Y., Yan, L., & Park, C. (2022). Multiple perspectives on analyzing risk factors in FMEA. Computers in Industry, 141, 103712.

Patle, D. S. , Agrawal, V. , Sharma, S. , & Rangaiah, G. P. . (2021). Plantwide control and process safety of formic acid process having a reactive dividing-wall column and three material recycles. Computers & Chemical Engineering, 147(2), 107248.

Pavlov, A. N., Abdurashitov, A. S., Koronovskii Jr, A. A., Pavlova, O. N., Semyachkina-Glushkovskaya, O. V., & Kurths, J. (2020). Detrended fluctuation analysis of cerebrovascular responses to abrupt changes in peripheral arterial pressure in rats. Communications in Nonlinear Science and Numerical Simulation, 85, 105232.

Pedrayes, O. D., Lema, D. G., Usamentiaga, R., & García, D. F. (2022). Detection and localization of fugitive emissions in industrial plants using surveillance cameras. Computers in Industry, 142, 103731.

Peng, L., Gao, D., & Bai, Y. (2021). A Study on Standardization of Security Evaluation Information for Chemical Processes Based on Deep Learning. Processes, 9(5), 832.

Qin, X., Zhou, Y., Guo, Y., Wu, D., Tian, Z., Jiang, N., ... & Wang, W. (2021, October). Mask is all you need: Rethinking mask r-cnn for dense and arbitrary-shaped scene text detection. In Proceedings of the 29th ACM International Conference on Multimedia (pp. 414-423).

Silva, V., Brzev, S., Scawthorn, C., Yepes, C., Dabbeek, J., & Crowley, H. (2022). A building classification system for multi-hazard risk assessment. International Journal of Disaster Risk Science, 13(2), 161-177.

Single, J. I., Schmidt, J., & Denecke, J. (2020). Knowledge acquisition from chemical accident databases using an ontology-based method and natural language processing. Safety Science, 129, 104747.

Standard, B., & IEC61882, B. S. (2001). Hazard and operability studies (HAZOP studies)-Application guide. International Electrotechnical Commission.

Stiernstroem, S. , Wik, O. , & Bendz, D. . (2016). Evaluation of frameworks for ecotoxicological hazard classification of waste. Waste Management, 58(dec.), 14-24.

Tanguy, L., Tulechki, N., Urieli, A., Hermann, E., & Raynal, C. (2016). Natural language processing for aviation safety reports: From classification to interactive analysis. Computers in Industry, 78, 80-95.

Tang, X., Yang, X., & Wu, F. (2020). Multifractal detrended fluctuation analysis parallel optimization strategy based on openMP for image processing. Neural Computing and Applications, 32(10), 5599-5608.

Tian, D., Li, M., Han, S., & Shen, Y. (2022). A Novel and Intelligent Safety-Hazard Classification Method with Syntactic and Semantic Features for Large-Scale Construction Projects. Journal of Construction Engineering and Management, 148(10), 04022109.

Wang, J., Shao, W., & Kim, J. (2020). Automated classification for brain MRIs based on 2D MF-DFA method. Fractals, 28(06), 2050109.

Wang, J., & Shao, W. (2021). Multifractal analysis with detrending weighted average algorithm of historical volatility. Fractals, 29(05), 2150193.

Wang, W. , Qin, J. , Yu, M. , Li, T. , & Li, J. . (2019). A reliability analysis of cfetr csmc heat treatment system based on RPN-HAZOP method. IEEE Transactions on Plasma Science, PP(99), 1-5.

Wang, Z., Zhang, B., & Gao, D. (2021). Text Mining of Hazard and Operability Analysis Reports Based on Active Learning. Processes, 9(7), 1178.

Wang(b), L., Zeng, X., Yang, H., Lv, X., Guo, F., Shi, Y., & Hanif, A. (2021). Investigation and application of fractal theory in cement-based materials: A review. Fractal and Fractional, 5(4), 247.

Wang, Z., Zhang, B., & Gao, D. (2022). A novel knowledge graph development for industry design: A case study on indirect coal liquefaction process. Computers in Industry, 139, 103647.

Wang(a) Z., Ming R., Gao, D., Zhuang L. (2022). Exploring industrial safety knowledge via Zipf law[J]. arXiv preprint arXiv: 2205.12636. https://arxiv.org/abs/2205.12636

Wang(c), Z., Liu, H., Liu, F., & Gao, D. (2022). Why KDAC? A general activation function for knowledge discovery. Neurocomputing, 501, 343-358.

Wen, T., & Cheong, K. H. (2021). The fractal dimension of complex networks: A review. Information Fusion, 73, 87-102.

Xu, X. Y., Wang, X. W., Chen, D. Y., Smith, C. M., & Jin, X. M. (2021). Quantum transport in fractal networks. Nature Photonics, 15(9), 703-710.

Yang, H., Zeng, B., Yang, J., Song, Y., & Xu, R. (2021). A multi-task learning model for chinese-oriented aspect polarity classification and aspect term extraction. Neurocomputing, 419, 344-356.

Zhang, R., Jia, C., & Wang, J. (2022). Text emotion classification system based on multifractal methods. Chaos, Solitons & Fractals, 156, 111867.

Zhao, Y., Zhang, B., & Gao, D. (2022). Construction of petrochemical knowledge graph based on deep learning. Journal of Loss Prevention in the Process Industries, 104736.

Zheng, Z., Lu, X. Z., Chen, K. Y., Zhou, Y. C., & Lin, J. R. (2022). Pretrained domain-specific language model for natural language processing tasks in the AEC domain. Computers in Industry, 142, 103733.

Zhong, B., Pan, X., Love, P. E., Sun, J., & Tao, C. (2020). Hazard analysis: A deep learning and text mining framework for accident prevention. Advanced Engineering Informatics, 46, 101152.

Zhu, X., Wu, J., Zhu, L., Guo, J., Yu, R., Boland, K., & Dietze, S. (2021). Exploring user historical semantic and sentiment preference for microblog sentiment classification. Neurocomputing, 464, 141-150.